\documentclass[letterpaper, 10 pt, conference]{ieeeconf}  

\IEEEoverridecommandlockouts                              

\usepackage[T1]{fontenc}
\usepackage{xcolor}
\usepackage{graphicx}
\usepackage{amsmath}
\usepackage{comment}
\usepackage{multirow}
\title{\LARGE \bf
SENSORIMOTOR GRAPH: Action-Conditioned Graph Neural Network for Learning Robotic Soft Hand Dynamics}

\author{João Damião Almeida$^{*\ 1}$, Paul Schydlo$^{*\ 1, 2}$, Atabak Dehban$^{1}$ and José Santos-Victor$^{1}$
\thanks{ $^*$Equal contribution $^{1}$Institute for Systems and Robotics, Instituto Superior Técnico, University of Lisbon $^{2}$Machine Learning Department, Carnegie Mellon University, USA}%
}

\begin{document}

\maketitle
\thispagestyle{empty}
\pagestyle{empty}

\begin{abstract}

\textcolor{black}{Soft robotics is a thriving branch of robotics which takes inspiration from nature and uses affordable flexible materials to design adaptable non-rigid robots. However, their flexible behavior makes these robots hard to model, which is essential for a precise actuation and for optimal control. For system modelling, learning-based approaches have demonstrated good results, yet they fail to consider the physical structure underlying the system as an inductive prior. In this work, we take inspiration from sensorimotor learning, and  apply a Graph Neural Network to the problem of modelling a non-rigid kinematic chain (\emph{i.e.} a robotic soft hand) taking advantage of two key properties: 1) the system is compositional, that is, it is composed of simple interacting parts connected by edges, 2) it is order invariant, \emph{i.e.} only the structure of the system is relevant for predicting future trajectories. We denote our model as the ``Sensorimotor Graph'' since it   learns   the system connectivity from observation and uses it for dynamics prediction.} 
\textcolor{black}{We validate our model in different scenarios and show that it outperforms the non-structured baselines in dynamics prediction while being more robust to configurational variations, tracking errors or node failures.}


\end{abstract}

\section{Introduction}

\subsection{Motivation}

\textcolor{black}{Soft robotics rose as a branch of robotics that uses soft materials, drawing inspiration from organic systems, to generate agents with a more flexible and adaptive behavior. These robots however are particularly challenging to model since: 1) they do not have a rigid structure, instead they obey to a continuous topology with infinite degrees of freedom, 2) the elastic
contact of soft gippers with the manipulated objects alters their shape and kinematics over time, and 3) imperfections that are inherent to some manufacturing techniques together with the time-varying parameters (\emph{e.g.} stiffness, friction, \emph{etc.}) caused by the mechanical and abrasive
wear of the rubbery materials \cite{georgethuruthelControlStrategiesSoft2018} call for self-calibration}.

\textcolor{black}{The need for self-calibration is not unique to soft robots. In fact,} during the first few months of life, newborns go through a phase of intensive sensorimotor learning, performing endless ``experiments'' with their own bodies, using motor babbling and observing the motor outcomes, to learn how to control their arms and hands \cite{Hofsten2004}.
This learning process to understand the structure and behaviour of our own body, called sensorimotor learning, is also observed when, for example, an amputee has to learn how to adapt to a prosthesis after a replacement surgery. 

\begin{figure}[h!]
    \centering
    \includegraphics[scale=0.44]{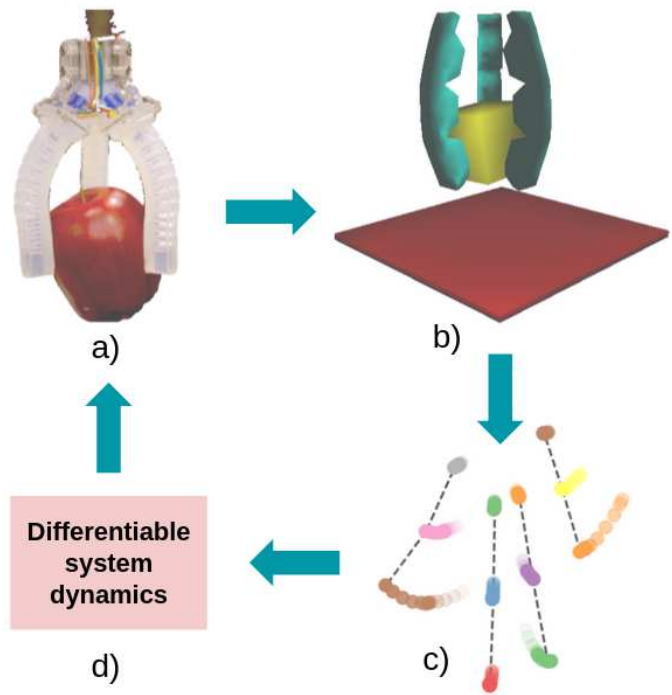}
    \caption{\textcolor{black}{Model based control applications require accurate prediction models. a) In this work we consider a Robotic Soft hand gripper system \cite{Homberg} b) simulate the gripper as soft material in a physics based simulator (SOFA) \cite{SOFA} under different conditions c) then given keypoints on the soft hand we learn a robust graph structured model d) the resulting differentiable system dynamics outperform the baseline models in a set of experiments validating different model assumptions \cite{neurips2018}.]}}
\end{figure}

Consider now an analogous situation for a robotic manipulator,  Fig. 1a), where a robot needs to understand how to use its end-effector. To effectively control this system we would first need to write down the equations which govern its motion. This approach is useful when the relations between the different parts are simple, but it becomes unfeasible for systems with complex non-trivial dynamics such as soft robotics and rubbery materials commonly used in prosthetic limbs. \textcolor{black}{In this work we explore a self-calibration strategy inspired by infant sensorimotor learning through motor babbling: by \textbf{observing the effects of actuation signals on a robotic soft-hand we infer the underlying structure in the form of a ``Sensorimotor Grap'' and predict future states conditioned on this graph}.}






\subsection{Related Work}

\subsubsection{\textbf{Forward Models}}

In classical approaches, the system model and parameters are assumed to be known in advance, and the controller design makes use of this information \cite{Pons99}. However, when the complexity of the system increases, considering the kinematics, under-actuation, and dynamics of dexterous hands,  it becomes very difficult to explicitly write precise and detailed model equations, which hampers the model-based control system design. Adaptive control emerged as a partial solution to this problem \cite{hsia}. By allowing parameters of the model to be estimated, it adapts to some levels of uncertainty regarding model parameters, such as e.g.: weight of objects. However, this approach is limited to a fixed structure of the model and is unsuitable for complex systems.






This has lead the community to consider learning based approaches to tackle the effects of under-modeling \cite{Ebert, Hafner2019}. \textcolor{black}{In the control domain these methods fall into two broad classes: model free and model based control \cite{polydorosSurveyModelBasedReinforcement2017}}. Model free variants of Reinforcement learning \cite{kober} can model and control a system by learning to map a system state to a control input, trained by optimizing a reward signal. Notwithstanding, reinforcement learning assumes the availability of interaction samples and a reward signal, that can be costly to collect. 

Model-based methods, on the other hand, decouples learning of the system dynamics and control of the system, reducing the need for costly interaction samples. Optimal control formulations based on differential models \cite{neurips2018, DPI}  optimize a cost function conditioned on the system states, reference states and actuation conditions.

Learning based approaches are generally based on two kinds of representations; a user-defined space such as 2D/3D position, or directly on the unprocessed sensory inputs, such as the image space. 
Image space representations take as input a sequence of frames and predict future frames, modelling the dynamics of the environment in pixel space by compressing  the information into a latent space \cite{Hafner2019}. This representation is not invariant to distribution shifts such as \textcolor{black}{environment}, lighting or camera pose changes. 

Methods based on explicit 3D representations on the other hand decouple the challenges related to lighting and camera poses. These approaches are based on either manually annotating the data, adding additional instrumentation in the form of markers, or learning to extract and identify key-point representations in an unsupervised manner \cite{NRI, Battaglia,DPI}. More structured representations have gained attention with the recent growing adoption of graph neural networks, which enable learning-based methods to exploit the structure inherent in the problem domain \cite{DPI, GNN5}, as we will see in the next section.

We focus on the question of how to learn to model a non-rigid kinematic chain, that is, a dynamical system of inter-connected parts\textcolor{black}{, learning a good model is key for model-based control applications}. We will assume explicit 3D point position representations, which allow us to limit the scope of this work to the problem of modelling the non-rigid kinematic chain of a robotic soft hand. Additionally, this representation is invariant to lighting and camera pose changes. In a real system the points can be obtained by either placing markers on the different fingers or by learning to segment a RGB-D camera stream into different finger segments.

\subsubsection{\textbf{Graph Neural Networks}}

Neural networks have enabled great progress in different areas of robotics and computer vision, \textcolor{black}{yet they are very sensitive to input distribution shifts such as changes to order of the input features \cite{GNN}. This is particularly relevant in situations where we expect tracking or sensing failures.} Recently, graph neural network models attempt to bridge the divide between classical structured and more recent deep learning data-driven approaches \cite{Battaglia}. The combination mitigates shortcomings by reasoning directly over explicit entities and their relations, the so called relational inductive bias. Imposing architectural constraints, this class of methods captures the inherent compositionality of the system, reasoning about the environment in terms of  entities and relations. This is postulated to be similar to how humans reason about their environment~\cite{Battaglia}. 

Graph neural networks are structured around a set of entities or nodes/vertices, $v_i$, and relations or edges between them, $e_k$. Together the nodes and edges form a graph that defines the dependencies among variables.

Information is propagated through a message passing mechanism in the form of successive edge and node update~\cite{GNN3}. Each edge is updated with the function $\phi^e$ accumulating and conditioning on the neighbouring nodes. Each node is updated with a function $\phi^v$ accumulating and conditioning on the surrounding edges. Successive iterations of these two functions converge towards the final node values. This mechanism is similar to the way information is collected in a convolutional kernel, where a pixel value is updated as a function of the information of the neighbouring pixels. Following the notation in \cite{Battaglia} this can be written as:
\begin{equation}
\begin{aligned}
\mathbf{e}_{k}^{\prime} &=\phi^{e}\left(\mathbf{e}_{k}, \mathbf{v}_{r_{k}}, \mathbf{v}_{s_{k}}\right) & \overline{\mathbf{e}}_{i}^{\prime} &=\rho^{e \rightarrow v}\left(E_{i}^{\prime}\right) \\
\mathbf{v}_{i}^{\prime} &=\phi^{v}\left(\overline{\mathbf{e}}_{i}^{\prime}, \mathbf{v}_{i}\right) & 
\end{aligned}
\end{equation}
Where $r_k$ is the receiving node and $s_k$ the sending node. $\mathbf{e}_{k}^{\prime} $, $\overline{\mathbf{e}}_{i}^{\prime}$, $\mathbf{v}_{i}^{\prime}$ and $N_e$ denote the new edge values, accumulated messages on the edge, new node values and number of edges respectively.  The messages are grouped in the following manner. 

\begin{equation}
\begin{gathered}
E_{i}^{\prime}=\left\{\left(\mathbf{e}_{k}^{\prime}, r_{k}, s_{k}\right)\right\}_{r_{k}=i, k=1 ; N^{e}} 
\end{gathered}
\end{equation}

Since the accumulation function, $\rho$, is order invariant, this kind of network is especially robust to changes in the order of the points, as long as the topology remains invariant. We take advantage of this property by exploiting the order invariance to make the model robust to possible re-identification or tracking issues. In other words this implies the network has no information about the node order, only the relation between them.

This structure is specially interesting for physical systems with structural constraints and has been applied to the problem domains of \textcolor{black}{non-rigid or articulated systems} \textcolor{black}{\cite{GNN5, IN, DPI}}. Here we continue this line of work and extend it to the domain of soft materials in robotic soft hands. 

GNN's have fueled a growing interest in learning models over graph structured data \cite{RNEM}, these methods generally send messages over a fully connected graph, not explicitly capturing the connectivity in the underlying system. In this work we follow the model proposed in \cite{NRI} which considers explicit system graph structure, that is, the connectivity of the graph mirrors the system connectivity.



\subsection{Our Approach} 

We propose to learn a structured differentiable action-conditioned dynamics model based on an explicit joint connectivity graph, the ``Sensorimotor Graph'', of a robotic soft hand. By observing \textcolor{black}{node position sequences (trajectories)} and actuation signals \textcolor{black}{(finger contractions)}, we infer the underlying system connectivity and then, to model the system dynamics, we condition our prediction model on the explicit connectivity graph structure. Imposing an explicit relational inductive bias on our system. 
 

 Model predictive control formulations can then take the learnt differentiable system dynamics model as constraints and optimize for an actuation signal that minimizes a control objective \cite{neurips2018}.

Learning dynamics in this manner enables us to decouple control and model learning. This implies we can learn the system in a self-supervised setting, by "babbling" motor \textcolor{black}{actuations}, that is, using a wide range of different control signals on the cable-driven actuator and observing the outcomes, avoiding costly reward samples. While we apply the methodology to the concrete example of a robotic soft hand, the method is general to any actuable system of interacting parts.


\subsection{Contributions}

We formulate the problem of learning a sensorimotor model for a robotic soft hand's non-rigid kinematic chain as learning an explicit connectivity graph which we designate Sensorimotor Graph. Additionally, we handled variable number of input nodes and propose the inclusion of explicit \textcolor{black}{actuation signals} to the graph forward model proposed in \cite{NRI}. This enables the down-stream application of the learnt model as a differentiable system dynamics model in an optimal control application. 

Furthermore, \textcolor{black}{we validate the proposed action-conditioned graph forward model by simulating robotic soft gripper data with varying parameters and configurations}, showing that it can capture the finger's coupled dynamics, and comparing it to different non-structured baseline models. Lastly we benchmark the model to understand robustness to conditions with increasing adversity.


\section{Problem Statement}

\textcolor{black}{In this work, we look at modelling a robotic soft hand to obtain a differential dynamics model. \textcolor{black}{We assume a hand composed of smaller parts (e.g. markers placed on multiple rubbery fingers) forming a non-rigid kinematic chain.} In a formal way, given $N_c$ nodes and their observations over time, $v^{1:t}_k$, $k\in[1,N_c]$, and actuation signals $u^{1:t+T}$, $u^{t}\in \mathcal{R}^{N_a}$, where $N_a$ is the number of actuation signals (e.g. one per finger), we wish to predict a sequence $v^{t+1:t+T}_k$ into the future. We model the system dynamics as the interaction of $N_c$ nodes represented as a "Sensorimotor Graph", this graph has nodes $V = \{v_k \in \mathcal{R}^n \}$, (representing e.g.:keypoints on a soft gripper), and a set of discrete edges, $z_{i,j}$ which represent the existence or non-existence of a relation between two nodes.} 

The system input vector, $x^t$ is given by the concatenation of the node state, $v_k^t$ \textcolor{black}{ with the finger contraction actuation signals $a^t$}. To be driven by external actuation signals, we consider the actions as part of the node input. This allows us to either map specific actions to specific nodes (providing the same action for the nodes belonging to the same finger), or provide all actions to all nodes and let the model infer the mapping between nodes and actuation signals. \textcolor{black}{We consider the more general case where the model infers the mapping between nodes and actions.} 



\textcolor{black}{In this work we propose to decouple the problem in two parts: 1) given node states, $v^{1:t}$ and actuation signals, $u^{1:t+T}$ infer the Sensorimotor Graph, $S$ , 2) then, given $S$, an initial state $v^t$, and the actuation sequence, $u^{t+1:t+T}$, predict a future sequence, $v^{t+1:t+T}$}. 






\section{Model}



\textcolor{black}{In this work we consider a framework for modelling and self-calibrating a robotic soft hand non-rigid kinematic chain. The model is inspired by two key ideas: 1) Sensorimotor babbling enables infants to understand their own kinematic structure and 2) we want to explicitly consider this physical structure, which we call the Sensorimotor Graph as an inductive bias in the prediction process. }

\textcolor{black}{For estimating this Sensorimotor Graph we integrate the Neural Relational Inference (NRI) model proposed in \cite{NRI}, which enables us to: 1) consider an \textbf{explicit} connectivity graph, 2) estimate this connectivity graph in an \textbf{unsupervised manner} while learning the system dynamics.}



\subsection{Structure}

The model is defined in two parts. A first part receives a sequence of system observations and infers the underlying edge structure, and a second part which takes this edge structure and system state, predicting the future state trajectory. We call these parts the encoder and decoder, respectively. Both these parts are trained jointly, either by providing ground truth labels for the edges or by learning them in an unsupervised manner end-to-end.

\subsection{Encoder}

\textcolor{black}{Following the notation in \cite{NRI}, the encoder takes a sequence of observed node trajectories and actuations to infer pair-wise interaction probabilities represented as edges $z_{ij}$}. 

\textcolor{black}{More formally, the encoder models the edge existence probability distribution, $q_\phi(z|x)$, as follows:}
\begin{equation}
    q_\phi(\mathbf{z}_{i,j}\vert \mathbf{x}) = \text{softmax} \left(f_{enc,\phi}(\mathbf{x})_{ij,1:N_c}\right)
\end{equation}

\textcolor{black}{Where \(f_{enc,\phi}(\mathbf{x})\) is a GNN acting on the fully-connected graph defined by all node pair permutations. The encoder computes iterative node-to-edge and edge-to-node message passing operations to simulate node interactions and then calculates the edge type. The sequential node and edge updates follow:}


\begin{equation}
\begin{aligned}
\mathbf{h}_{j}^{1} &=f_{\text {emb }}\left(\mathbf{x}_{j}\right) \\
v \rightarrow e: \quad\mathbf{h}_{(i, j)}^{1}&=f_{e}^{1}\left(\left[\mathbf{h}_{i}^{1}, \mathbf{h}_{j}^{1}\right]\right) \\
e \rightarrow v: \, \quad\quad \mathbf{h}_{j}^{2}&=f_{v}^{1}\left(\sum_{i \neq j} \mathbf{h}_{(i, j)}^{1}\right) \\
v \rightarrow e: \quad \mathbf{h}_{(i, j)}^{2} &=f_{e}^{2}\left(\left[\mathbf{h}_{i}^{2}, \mathbf{h}_{j}^{2}\right]\right)
\end{aligned}
\end{equation}

More specifically, in the first pass the node to edge messages are computed using an embedding function, $f_{emb}$. Embedded node values from nodes $i$ and $j$ are then used to update the edge $e_{i,j}$ between them. After updating the edge state, another embedding function, $f_v$, takes the edge value and uses it to compute edge-to-node messages. The edge-to-node messages are aggregated for all edges connected to node $i$. Then, a final node-to-edge message passing operation infers the final edge value from which the edge existence probability is calculated using a softmax function. 

\begin{equation}
\mathbf{z}_{i j}=\text{softmax}\left(\left(\mathbf{h}_{(i, j)}^{2}+\mathbf{g}\right) / \tau\right)
\end{equation}

Given the distribution, it is possible to either train the model in a supervised setting by providing ground truth labels for the edges, or training the model end-to-end by sampling edges from the inferred distribution and using these as graph structure for the subsequent trajectory prediction step. The NRI model uses a continuous approximation of the discrete distribution to effectively sample from a discrete distribution. Here, \textcolor{black}{$g\in R^K$ is a vector of i.i.d. samples drawn from a Gumbel(0,1) distribution} and $\tau$ is a temperature parameter that adjusts the "softness" of the resulting distribution (when $\tau \rightarrow 0$, the distribution converges to one-hot samples, as described in \cite{NRI}). 

\subsection{Decoder}

The second part, called the decoder, takes the edges and uses them as explicit structure for the graph computations in the trajectory prediction step. Just like the encoder, the trajectory predictions are based on a sequence of node-to-edge and edge-to-node passes. An initial accumulation function updates the edge value by weighting adjacent nodes by the edge existence probability, \textcolor{black}{where} $z_{i,j,k}$ represents the k-th entry of the vector $z_{i,j}$ and $\tilde{f}_e^k$ is the embedding function associated to this edge type $k$. \textcolor{black}{Then,} edges are aggregated in the nodes through an order invariant summation and transformed by an embedding function, $\tilde{f}_v$. \textcolor{black}{Adding them to} the current node state \textcolor{black}{outputs} the mean value, $\mu_j^{t+1}$ for the normal distribution from which the next node state, $x_j^{t+1}$ is sampled, \textcolor{black}{with variance $\sigma^2$}. 

\begin{equation}
\begin{aligned}
v \rightarrow e: \quad \tilde{\mathbf{h}}_{(i, j)}^{t} &=\sum_{k} z_{i j, k} \tilde{f}_{e}^{k}\left(\left[\mathbf{x}_{i}^{t}, \mathbf{x}_{j}^{t}\right]\right) \\
e \rightarrow v: \quad\,\boldsymbol{\mu}_{j}^{t+1}&=\mathbf{x}_{j}^{t}+\tilde{f}_{v}\left(\sum_{i \neq j} \overline{\mathbf{h}}_{(i, j)}^{t}\right) \\
p\left(\mathbf{x}_{j}^{t+1} \mid \mathbf{x}^{t}, \mathbf{z}\right) &=\mathcal{N}\left(\boldsymbol{\mu}_{j}^{t+1}, \sigma^{2} \mathbf{I}\right)
\end{aligned}
\end{equation}

The decoder can be structured as a transformation function modelled as an MLP or as an RNN, acting recursively on its own internal state when updating the node using a memory cell.



\subsection{Loss}

Following the formulation in \cite{NRI}, the model distributions are formalized and trained as a Variational Auto-Encoder. We define a Evidence Lower Bound loss in the following way. 

\begin{equation}
\begin{split}
\mathcal{L}=\mathbf{E}_{q_{\phi}(\mathbf{z} \mid \mathbf{x^{0:t}})}\left[\log p_{\theta}(\mathbf{x^{t+1:T}} \mid \mathbf{z})\right]\\-\mathrm{KL}\left[q_{\phi}(\mathbf{z} \mid \mathbf{x^{0:t}}) \| p_{\theta}(\mathbf{z})\right]
\end{split}
\end{equation}

The first term term measures the expected loss weighted by the edge type distribution. This weights predicted trajectories, x, given by the decoder probability $p_\theta(x^{t+1:T}|z)$, which are conditioned on improbable edge configurations, $q_\phi(z|x^{0:t})$, less than the trajectories which are predicted from very probable edge configurations, given the observation of a past trajectory. The second term encourages a low Kullback–Leibler divergence between the the edge type distribution and a prior, soft constraining the distribution to be close to the prior. Both the terms together enable the model to learn edge distributions in a supervised manner, given ground truth edge labels, or unsupervised, without any labels, when training with future state predictions simultaneously. We refer the reader to \cite{NRI} for a formal derivation of the loss terms.

\section{Experimental Setup}

\subsection{Data Collection}

\begin{figure}[!ht]
    \centering
    \includegraphics[scale=0.23]{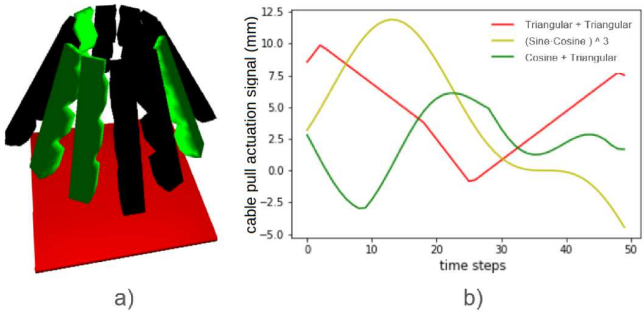}
    \caption{Sources of variability in the datasets: a) finger positions in a dodecagon arrangement with one configuration selected (green), configurations are generated by permutations of the selected fingers b) Sample  of cable pull actuation signals used in the Trainset .}
    \label{simulator}
\end{figure}

In order to collect diverse data from a soft hand gripper in motion, the Simulation Open Framework Architecture (SOFA) \cite{SOFA} is used, alongside with the Soft Robotics Toolkit \cite{Soft}. \textcolor{black}{The system consists of a cable-actuated hand, where each finger with three soft phalanges is actuated by a pulling cable that contracts or distends.} The finger base is fixed and the variation in the cable pull actuation signal allows the soft finger to experience a wide range of motion. Variability is added to the scene by including fingers in different number and configurations around a dodecagon (Fig. \ref{simulator}a)). The actuation signal (Fig. \ref{simulator}b)) and elasticity of each finger vary in different trials. Each run in the SOFA environment creates a sequence of 100 sampled time steps sampled at a frequency of 8 samples per second.  For each of the four fingers, three points are sampled, one for each \textcolor{black}{phalanx, in a total of twelve nodes.}

\subsection{Training and Validation Sets}

A total of six different data sets are extracted from the soft gripper simulation. The training set for all tests includes only four fingers in a total of thirty different configurations (relative positions) and with four different values of elasticity \textcolor{black}{(Poisson ratio and Young modulus)}. We will reference this set as Trainset. For each sample, the type of motion applied to the cable pull actuation signal for every finger is the same, although the characteristics of the motion - amplitude, frequency and phase - are different and randomized. Three different types of motion are used (Fig. \ref{simulator}b). Our first set, TestBase, includes novel configurations, some combinations of motion and elasticity unseen at training time, as well as a different cable pull actuation signal. 

We created four sets to test the model in different validation sets with \textcolor{black}{single} varying conditions. The validation sets are setup to understand: 1) TestMotion: How well the model generalizes to previously unseen motions, 2) TestConfig: Effect of system configuration (relative position of fingers) deviating from the training configurations, either due to sensor measurement errors or due to changes in the end-effector, 3) TestFingers: Adaptation to changes in the number of fingers, either due to mechanical failure of one of the joints or sensor measurements failing to provide the position for any of the fingers or points, 4) TestShuffle: Robustness to changes in the order of the joint measurements, either due to tracking errors or wrong correspondences between detected parts.

More specifically the four validation sets are used to test the models generalization to motion, relative position, number of fingers and order of the points, respectively. The first test set, TestMotion, uses three different motions (resulting from operations between trigonometric functions) and includes random noise. The second test set, TestConfig, uses the same motions and elasticities as the training set but for thirteen unseen configurations. The third test set, TestFingers, uses only three-finger configurations with the same motions and elasticities whereas the fourth, TestShuffle, and last validation set, shuffles the order of the twelve nodes for each sample.
\begin{table}[]
\caption{ \textcolor{black}{Description of validation sets} }
\centering
\bgroup
\def\arraystretch{1.5}%
\begin{tabular}{|l|p{0.7\linewidth}|}
\hline
\textbf{Test Sets} & \textbf{Description} \\ \hline
TestBase &  \textbf{Overall} generalization: validation uses new combinations of motions, elasticities and configurations  \\ \hline
TestMotion &  Generalization to \textbf{unseen motions}: validation uses comnbinations of 3 different trigonometric functions for motion (and includes noise)  \\ \hline
TestConfig & Generalization to \textbf{unseen configurations}: validation uses 13 unseen finger configurations   \\ \hline
TestFinger & Generalization to \textbf{different number of nodes}: validation uses three- instead of four-finger gripper.   \\ \hline
TestShuffle & Generalization to \textbf{new order of points}: validation data is the same as training but node order is shuffled. \\ \hline
\end{tabular}
\egroup
\end{table}

\subsection{Experiments}

The model is evaluated in two separate sets of experiments: 1) comparison to other methods proposed in the literature and 2) ablation studies to validate the robustness of the proposed approach under different conditions. 

The first part looks at our approach in relation to baseline models. In this set of experiments the following models are quantitatively compared: \textcolor{black}{1) Non-learning method} 2) Linear, 3) Multi-layer Perceptron (MLP), 4) Long Short Term Memory sequence model (LSTM), 5) Neural Relational Inference (NRI). \textcolor{black}{In the non-learning method, the position remains the same as the last known point for the whole prediction time.} We consider the linear and \textcolor{black}{MLP} multi-layer perceptron models as they validate if the task can be solved by simple linear or non-linear interpolation of the trajectory points. Long Short Term Memory model was chosen as being representative of a non-structured model, enabling us to compare the performance of a structured sequence model with a non-structured sequence model. 

The second part looks more closely at how different conditions affect the model under analysis \textcolor{black}{(ablation study)}. For this purpose, only the most relevant baselines are considered, alongside a structured model.




\section{Results and Discussion}

\subsection{Setup}

The experiments are structured around two parts. The first, is a quantitative comparison of how the different models perform over the first validation set, TestBase. The second, evaluates how each source of variability affects the different models and how they react to different types of generalization

\subsection{Model Comparison}

For the first experiment, the structure-based approach (NRI) is compared to three different baselines: Linear, MLP and LSTM. The NRI model has two model variations on the decoder part of the model, it can either be based on a MLP (NRI-mlp) or an RNN structure (NRI-rnn), in this experiment we consider both variants. For each of the possible decoder structures the encoder has two additional variations: supervised and unsupervised. The supervised encoder receives ground truth edges for the connections between the finger nodes \textcolor{black}{during training} where as the unsupervised learns the connections in an end-to-end manner while learning to predict future trajectories. 

\renewcommand{\arraystretch}{1.5}

\begin{table}[h!]
\centering
\caption{Mean Squared Error for model and baselines}
\begin{tabular}{lccccl} 
\hline
 \textbf{Model} & \textbf{$\text{MSE}_5$[x$10^{\text{-}4}$]} & \textbf{$\text{MSE}_{10}$[x$10^{\text{-}3}$]} & \textbf{$\text{MSEn}_{10}$[x$10^{\text{-}2}$]}\\  
 \hline
 last position & 6.597 & 1.656 & 2.333 \\
 \hline
 Linear & 7.987 & 1.268 & 1.839 \\ 
 MLP & 7.878 & 1.117  & 1.698 \\
 LSTM & 6.157 & 1.354 & 1.964 \\
 \hline
 uns. NRI-mlp & 4.962 & 1.336  & 1.926 \\
 uns. NRI-rnn & \textbf{2.200} & \textbf{0.801} & \textbf{1.142}  \\  
 \hline
 sup. NRI-mlp & 4.753 & 1.281  & 1.846 \\
 sup. NRI-rnn & \textbf{2.108} & \textbf{0.746}  & \textbf{1.063} \\
 \hline
\end{tabular}
\label{table:1}
\end{table}

In this experiment, all models are trained for 10 predictions steps. In the variations of the NRI with no ground-truth edge information, a non-overlapping set of fifty time steps is provided to the encoder to estimate the connectivity graph. For all the baselines and for the NRI-rnn, there is an initialization or "burn-in" phase where the models have access to the five time steps before the sequence that is considered for the evaluation. The LSTM "burn-in" phase, similarly to the NRI, has fifty steps before the ten step prediction window. This allows the LSTM to initialize its internal state.

Table \ref{table:1} shows how the proposed model compares to different baseline models\textcolor{black}{, where the Mean Squared Error (MSE) represents the deviation between the prediction and the real trajectory and MSEn stands for the MSE normalized to the travelled distance (MSEn=1 means that the prediction of one point has deviated as much as the distance the point travelled). The values in the tables may appear low, but this is due to the node positions being normalized with respect to the data-set statistics. Therefore, this MSE has no direct physical meaning, this metric allows us to capture the quantitive performance difference between the models. We add a non-learning model to provide a baseline for the error when the predicted trajectory stands still.} Here we can see that the proposed model (NRI) outperforms the baseline models in the first testset, TestBase, which has a diverse set of motions, configurations and elasticities, when the prediction step equals the training step. Within the different variations of the NRI model we find that the best combination is the supervised variant (sup. NRI-rnn). We will look more closely at how this NRI model variant performs under different conditions in the next experiment. 

\begin{figure}[h!]
    \centering
    \includegraphics[scale=1.05]{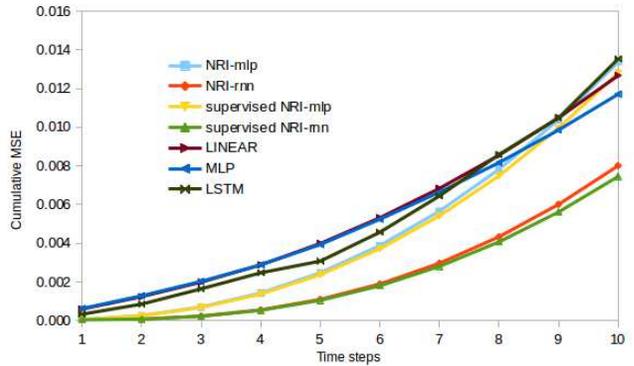}
    \caption{Cumulative mean square error (MSE) over 10 time-steps for different models. }
    \label{temporal_10}
\end{figure}

Figure \ref{temporal_10} shows how the models compare when predicting for 10 steps ahead, the same number they were trained on. This validation set includes small variations in motion, elasticity and configurations. We can see that the NRI model using an RNN decoder outperforms the other variations considerably. We notice that while the NRI-rnn has better performance for the first steps, particularly relevant in planning, the accumulated error starts deteriorating as we predict beyond the training distribution, further into the future. We will look at this more closely in the ablation studies. 

\subsection{Ablation Study}

We will now look at the second set of experiments (Tab. \ref{table:ablation}) which seek to see how the model performs under different input distribution shifts. Models are all trained for 10 steps and test sets are 25 steps long. The first testset, TestBase, establishes a baseline for how well models perform in auto-regressively predicting more steps ahead than they were trained on. The four test-sets introduced in the experimental setup all look at different variations on the input data. Refer to Table 1 for an overview of the different test-sets. This enables us to study the effects of the different sources of prediction error separately. For this ablation study we will consider the best performing NRI model (NRI-rnn) and two relevant baselines, LSTM and MLP.

\begin{table}[h!]
\centering
\caption{\textcolor{black}{Ablation study for different input conditions}}
\begin{tabular}{llllll}
\hline
{[}x$10^{\text{-}3}${]} & TestBase & TestMot. & TestConf. & TestFing.                   & TestShuff.                  \\ \hline
MLP                           & \textbf{2.250}     & \textbf{1.051}     & 3.090     & - & - \\
LSTM                          & 4.762     & 6.786     & 3.015     & 2.843                        & 3.323                        \\
NRI                           & 3.303     & 4.482     & \textbf{2.056}     & \textbf{1.469}                        & \textbf{1.502}                        \\ \hline
\end{tabular}
\label{table:ablation}
\end{table}

From the initial baseline on TestBase, we can see that the NRI performs worse than the baseline models when generalizing to the distribution shift introduced by predicting longer sequences. In this first Testset, the MLP seems to be capable of predicting longer sequences with a lower mean error. The model also proves to have good generalization to new motions. This advantage of the MLP compared to the structure-based models can be justified with the fact that it has fewer model parameters and is able to exploit the relations between the nodes in adjacent positions of the input vector. The NRI, due to its order invariant graph-based structure, only has access to the nodes without explicit order like in a vector-based representation of the nodes. 

This, however, makes the MLP quite fragile to changes or disturbances in the data. Testing the same model in a situation with new relative positions of the fingers, TestConfig, MLP's prediction error increases considerably. Under the same conditions, we can see the NRI outperforming the baseline models. Under the TestFingers dataset, the MLP is not able to generalize and has an error significantly higher than the other models and is therefore not considered in this validation set. We can again see the NRI outperforming the baseline models in this condition. The last testset, TestShuffled, considers the situation where points are shuffled and the points at test are in a different order than the configurations it was trained on. This models a tracking or re-identification error. Here, again, we can see that, due to its order invariance properties, the NRI model outperforms the baseline models.

\subsection{\textcolor{black}{Connectivity}}

\textcolor{black}{
The model can be trained both in an unsupervised and supervised manner. In this section, we look at how well the model performs in recovering the edges when trained in both these regimes. For this, we trained the model with and without labels and measured the accuracy of the edge predictions given node trajectories against a held out ground-truth label set. In Table IV we see that: 1) connectivity inference contributes positively to the dynamics prediction, 2) supervised training recovers g.t. connectivity and outperforms the unsupervised variant, 3) unsupervised training does not usually recover ground-truth connectivity. This last phenomena happens since: 3.1) unsupervised edge estimation approach does not see any labels for the edge types and might arrive at a similar, but opposite, correspondence, 3.2) intricate counter-intuitive relations in the graph emerge, particularly finger-to-finger connections, similar to what was pointed out in \cite{NRI,structuralRNN}. A future direction to encourage the model to infer the real physical connectivity could be to impose sparsity constraints on the edges and consider the possibility of curriculum learning by exploring the finger synergies from a more coarse to a fine level.
}

%
%
%
%

\begin{table}[h!]
\centering
\caption{Connectivity inference performance}
\begin{tabular}{ccll}
\hline
\multicolumn{4}{c}{\textbf{NRI-rnn}}                                      \\ \hline
\textbf{training} & \textbf{$\text{MSE}_{10}$} [x10\(^{\text{-}4}\)] & \textbf{accuracy} & \textbf{F1score} \\
supervised        & 7.499    & 0.998 & 0.991 \\
unsupervised      & 8.062    & 0.486 & 0.176 \\ \hline
\end{tabular} \label{tab:connectivity}
\end{table}

Moreover, we test increasing the number of edge types from 2 to 3 (for the supervised variant we encode an extra relation type as the opposite direction of the existing edge). In Table V we notice the performance of both the supervised and unsupervised NRI-rnn marginally improve when we have this additional degree of freedom.


\begin{table}[h!]
\centering
\caption{Influence of the number of edge types parameter in the performance of the decoder}
\begin{tabular}{ccc} 
\hline
\multicolumn{3}{c}{\textbf{NRI-rnn} [x10\(^{\text{-}4}\)]}            \\ \hline
\textbf{ d } & \textbf{training} & \(\mathbf{MSE_{10}}\) \\ \hline
\multirow{2}{*}{2}  & supervised    & 8.045   \\  & unsupervised      & 8.206 \\ \hline 
\multirow{2}{*}{3} & supervised    & \textbf{7.588}     \\
   & unsupervised      & \textbf{8.083}     \\ \hline
\end{tabular} \label{tab:edgetypes}
\end{table}

\subsection{\textcolor{black}{Prediction step}}

\textcolor{black}{Finally, we consider training the Sensorimotor Graph model for different prediction sizes. This hyper-parameter sets the size of the prediction length in training and test. The prediction step should be big enough to contribute to a relevant forecast (used in model-predictive control) but not too large as this may result in unstable and chaotic dynamics. All models are trained with teacher forcing and predict the positions auto-regressively. When training the model for larger sequences, as we can see in Table IV, the model performed better at test time as well. Moreover, it is also noticeable that increasing the prediction step does not compromise accuracy in the initial time steps: larger windows benefit predictions right from the start. The comparative inference speed of the NRI model is depicted in Table VII. For comparison, all models were benchmarked on a single core of a i7 2.6GHz CPU.}

\begin{table}[h!]
\centering
\caption{Influence of prediction length parameter during training in prediction performance}
\begin{tabular}{llllll}
\hline
[$10^{\text{-}3}$] & $\text{PS}_5$      & $\text{PS}_{10}$  & $\text{PS}_{15}$      & $\text{PS}_{20}$     & $\text{PS}_{25}$     \\ \hline
$\text{MSE}_5$       & 0.115 & 0.114 & 0.110 & 0.099 & 0.097 \\
$\text{MSE}_{10}$      & 0.840 & 0.825 & 0.785 & 0.688 & 0.647 \\
$\text{MSE}_{15}$      & 2.498 & 2.433 & 2.291 & 1.935 & 1.768\\ 
$\text{MSE}_{20}$      & 5.307 & 5.121 & 4.785 & 3.928 & 3.501 \\
$\text{MSE}_{25}$      & 9.020 & 8.615 & 8.008 & 6.518 & 5.602 \\ \hline
\end{tabular}
\end{table}

\begin{table}[h!]
\centering
\caption{Inference speed of different models}
\begin{tabular}{lllll}
\hline
Model                                                         & MLP   & NRI-mlp & NRI-rnn & LSTM \\ \hline
\begin{tabular}[c]{@{}l@{}}Iteration\\ time (ms)\end{tabular} & 0.164 & 99.8    & 643     & 666  \\ \hline
\end{tabular}
\end{table}

\subsection{Discussion}

We evaluated the model against multiple baseline models in different sets of experiments. In the first experiment (Fig \ref{temporal_10} we found that the proposed action-conditioned NRI model outperforms the baseline models on TestBase with a test time prediction step of 10. This data-set tests for small variations in system attributes and validates the robustness of the proposed model. Furthermore, from this first set of experiments we can see that the supervised RNN variant of the NRI model outperforms the other variants.


In the second set of experiments, we found that the NRI model performs worse than the baseline models when tested to auto-regressively predict longer sequences on a TestBase testset of lengh 25. When tested under different configurations or number of nodes, it outperforms the baseline models, validating the robustness to sensor noise. Additionally, when tested on a test-set of shuffled points, it shows it is able to reconstruct the connectivity graph and outperform the baseline models, validating robustness to re-identification or tracking issues.




\section{ Conclusions and Future Work}

In this work we propose applying a GNN - the ``Sensorimotor Graph''model - to the domain of modelling a robotic soft hand. More specifically, we described the underlying compositional assumptions of the system and showed how a structured GNN successfully satisfies them. We denote our model as the ``Sensorimotor Graph'' as it is inspired on the way human infants acquire sophisticated motor capabilities (e.g. of the human hand) through sensorimotor learning and motor babbling. We chose the Neural Relation Inference as a candidate GNN for the problem domain. We showed how the model's explicit graph representation enables us to exploit the underlying physical structure of the system. To evaluate the model, we benchmarked it against non-structured baselines and evaluated the model robustness to gradually more adverse conditions. 

We found that the model performs well under conditions which require generalizing to changes in structure. More specifically, it performed well under conditions where the number and location of the fingers changed. Furthermore, we looked at different variations of the model and observed that out of two model variations on the encoder architecture, supervised and unsupervised, the supervised model performs better.



We showed that GNNs, more specifically the Neural Relational Inference model, can be used to robustly model a system of interacting soft material parts in the form of a differentiable dynamics model. This opens up promising avenues for combining the learnt model with a non-convex optimal control framework, using the differentiable dynamics model as systems constraints. 


We believe that the encouraging results and generalization capabilities of our ``Sensorimotor Graph'' model pave the way to the ability to acquire the structure and dynamics of complex systems and will be of utmost importance for the deployment of complex non-rigid kinematic chains and affordable articulated end-effectors such as soft hands for advanced manipulation.

\addtolength{\textheight}{-8cm}   

\section*{ACKNOWLEDGMENT}

This work was supported by a doctoral grant from FCT (SFRH/BD/139092/2018).

{
\bibliographystyle{IEEEtran}
\bibliography{IEEEexample}
}

\end{document}